# Superhuman performance of a large language model on the reasoning tasks of a physician

Peter G. Brodeur[1]*, Thomas A. Buckley[2]*, Zahir Kanjee[1], Ethan Goh[3,4], Evelyn Bin Ling[5], Priyank Jain[6], Stephanie Cabral[1], Raja-Elie Abdulnour[7], Adrian D. Haimovich[8], Jason A. Freed[9], Andrew Olson[10], Daniel J. Morgan[11,12], Jason Hom[5], Robert Gallo[13], Liam G. McCoy[1,14,15], Haadi Mombini[16], Christopher Lucas[1], Misha Fotoohi[1], Matthew Gwiazdon[1], Daniele Restifo[1], Daniel Restrepo[17], Eric Horvitz[18, 19], Jonathan Chen[3,4,5]**, Arjun K. Manrai[2]**, Adam Rodman[1]**

*co-first authors
**co-senior authors

Correspondence: Arjun_Manrai@hms.harvard.edu / arodman@bidmc.harvard.edu

1. Department of Internal Medicine, Beth Israel Deaconess Medical Center, Boston, Massachusetts
2. Department of Biomedical Informatics, Harvard Medical School, Boston, Massachusetts
3. Stanford Center for Biomedical Informatics Research, Stanford University, Stanford, California
4. Stanford Clinical Excellence Research Center, Stanford University, Stanford, California.
5. Department of Internal Medicine, Stanford University School of Medicine, Stanford, California
6. Department of Internal Medicine, Cambridge Health Alliance, Cambridge, Massachusetts
7. Division of Pulmonary and Critical Care Medicine, Brigham and Women's Hospital, Boston, Massachusetts
8. Department of Emergency Medicine, Beth Israel Deaconess Medical Center, Boston, Massachusetts
9. Department of Hematology-Oncology, Beth Israel Deaconess Medical Center, Boston, Massachusetts
10. Department of Hospital Medicine, University of Minnesota Medical School, Minneapolis
11. Department of Epidemiology and Public Health, University of Maryland School of Medicine, Baltimore, Maryland
12. Veterans Affairs Maryland Healthcare System, Baltimore, Maryland
13. Center for Innovation to Implementation, VA Palo Alto Health Care System, Palo Alto, California
14. Institute for Medical Engineering and Science, Massachusetts Institute of Technology, Cambridge, Massachusetts
15. Division of Neurology, University of Alberta, Edmonton, AB, Canada
16. Technology and Innovation Group, Beth Israel Lahey Health, Boston, Massachusetts
17. Department of Medicine, Massachusetts General Hospital, Boston, Massachusetts
18. Microsoft Corp, Redmond, Washington
19. Stanford Institute for Human-Centered Artificial Intelligence, Stanford, California

## ABSTRACT

A seminal paper published by Ledley and Lusted in 1959 introduced complex clinical diagnostic reasoning cases as the gold standard for the evaluation of expert medical computing systems, a standard that has held ever since. Here, we report the results of a physician evaluation of a large language model (LLM) on challenging clinical cases against a baseline of hundreds of physicians. We conduct five experiments to measure clinical reasoning across differential diagnosis generation, display of diagnostic reasoning, triage differential diagnosis, probabilistic reasoning, and management reasoning, all adjudicated by physician experts with validated psychometrics. We then report a real-world study comparing human expert and AI second opinions in randomly-selected patients in the emergency room of a major tertiary academic medical center in Boston, MA. We compared LLMs and board-certified physicians at three predefined diagnostic touchpoints: triage in the emergency room, initial evaluation by a physician, and admission to the hospital or intensive care unit. In all experiments—both vignettes and emergency room second opinions—the LLM displayed superhuman diagnostic and reasoning abilities, as well as continued improvement from prior generations of AI clinical decision support. Our study suggests that LLMs have achieved superhuman performance on general medical diagnostic and management reasoning, fulfilling the vision put forth by Ledley and Lusted, and motivating the urgent need for prospective trials.

## INTRODUCTION

Artificial intelligence (AI) diagnostic support tools have been studied since the 1950s, following a landmark paper published in *Science* by Ledley and Lusted (*1*) who advocated for case-based benchmarks as an evaluation standard, a standard that has held for over the past half century (*1–6*). In particular, the *New England Journal of Medicine* clinicopathological case conference series has been seen an aspirational goal post, tested by every differential diagnosis generator from primitive Bayesian systems, symbolic rules-based systems, and natural-language symptom checkers. Recently, large language models (LLMs) have consistently outperformed older models on these challenging cases, mirroring their performance in professional licensing exams, mathematics questions, software engineering, and engineering problems (*7–10*).

However, recent studies of LLMs in medicine have focused on narrow diagnostic tasks or on curated and constrained clinical vignettes (*11–13*). More importantly, most studies of LLMs for diagnosis and management to date have lacked human physician baselines. This was justifiable in previous generations of technology because of the overall poor performance on benchmarks. Given rapid improvement in LLMs and increasing “benchmark saturation,” it is necessary to establish human baselines and study clinically grounded tasks. Here, we comprehensively evaluate the diagnostic and management reasoning capabilities of an advanced LLM (OpenAI o1 series) against a baseline of hundreds of physicians. We further study LLM second opinions in a blinded fashion against an expert physician baseline on randomly-selected patients in a major academic tertiary care emergency department in Boston, MA.

## RESULTS

### Quality of Differential Diagnoses on New England Journal of Medicine Clinicopathological Conferences

We first evaluated o1-preview using the clinicopathologic conferences (CPCs) published by the *New England Journal of Medicine* (NEJM), a standard for the evaluation of differential generators since the 1950s (*1*). There was substantial agreement between the two physicians evaluating the quality of o1-preview's differential diagnosis (agreement on 120/143 cases [84%], $\kappa$=0.66). o1-preview included the correct diagnosis in its differential in 78.3% of cases (95% CI, 70.7% to 84.8%) (Figure 1). The first diagnosis suggested was the correct diagnosis in 52% of cases (95% CI, 44% to 61%). We did not find evidence of a significant difference in performance before and after the pre-training cutoff date for o1-preview (79.8% accuracy before, 73.5% accuracy after, p=0.59). Examples of o1-preview solving a complex case are shown in Table 1.

On 70 cases evaluated using GPT-4 in a prior study (*12*), o1-preview produced a response with the exact or a very close diagnosis in 88.6% of cases, compared to 72.9% of cases by GPT-4 (p=.015, Figure 2).

We next evaluated the ability of o1-preview to select the next diagnostic test in the NEJM CPCs. Two physicians scored the suggested test plan produced by o1-preview (agreement on 113/132 cases [86%], $\kappa$=0.28), with respect to the actual management of the patient described in the CPC. The proportion of agreements was high, but the kappa was low due to severe class imbalance. In 87.5% of cases, o1-preview selected the correct test to order, in another 11% of cases the chosen testing plan was judged by the two physicians to be helpful, and in 1.5% of cases it would have been unhelpful (Figure 3). Examples are shown in Table 2.

### Presentation of reasoning in NEJM Healer Diagnostic Cases

We used 20 clinical medical cases from the NEJM Healer curriculum (*14*) that were also evaluated in a prior study using GPT-4 (*15*). NEJM Healer cases are virtual patient encounters designed for the assessment of clinical reasoning (*14*). There was substantial agreement of Revised-IDEA (R-IDEA) scores, a validated 10-point scale for evaluating four core domains of documenting clinical reasoning (*16*), between the two physicians (agreement on 79/80 [99%] cases, $\kappa$=0.89). For 78/80 of the cases, o1-preview achieved a perfect R-IDEA score. This significantly outperformed GPT-4 (47/80, $p<0.0001$), attending physicians (28/80, $p<0.0001$), and resident physicians (16/80, $p<0.0001$) as shown in Figure 4A. We measured the proportion of "cannot-miss" diagnoses identified by o1-preview during the initial triage presentation (Figure 4B). The median proportion of "cannot-miss" diagnoses included for o1-preview was 0.92 (IQR, 0.62 to 1.0) though this was not significantly higher than GPT-4, attending physicians, or residents.

### *Grey Matters* Management Cases

We used five clinical vignettes based on real cases from a previous study developed and scored with consensus methods from 25 physician experts (*17*). Each clinical vignette was presented to the model and was followed by a series of questions regarding next steps in management. Two physicians scored responses by o1-preview for the five cases, with substantial agreement (κ=0.71). The median score for the o1-preview per case was 86% (IQR, 82%-87%) (Figure 5A) as compared to GPT-4 (median 42%, IQR 33%-52%), physicians with access to GPT-4 (median 41%, IQR 31%-54%), and physicians with conventional resources (median 34%, IQR 23%-48%). Using the mixed-effects model, o1-preview scored 41.6 percentage points higher than GPT-4 alone (95% CI, 22.9% to 60.4%; $p < 0.001$), 42.5 percentage points higher than physicians with GPT-4 (95% CI, 25.2% to 59.8%; $p < 0.001$), and 49.0 percentage points higher than physicians with conventional resources (95% CI, 31.7% to 66.3%; $p < 0.001$).

### Landmark Diagnostic Cases

We used six clinical vignettes from a previous study that compared GPT-4 to 50 generalist physicians (*18*). The cases derive from a landmark study of computer-based diagnostic systems, containing the history of present illness, past medical history, physical exam, and diagnostic studies (*19*). The cases have never been publicly released specifically to protect evaluation validity against memorization. Two physicians scored responses by o1-preview to the six diagnostic reasoning cases, with moderate agreement for total score (κ=0.42). The median score for the o1-preview model per case was 97% (IQR, 95%-100%) (Figure 5B). This is compared to historical control data where GPT-4 scored 92%, (IQR 82%-97%), physicians with access to GPT-4 scored 76%, (IQR 66%-87%), and physicians with conventional resources (median 74%, IQR 63%-84%). Using the mixed-effects model, o1-preview performed comparably to GPT-4 (4.4% higher, 95% CI, -19.0% to 27.7%; $p = 0.7$), physicians with GPT-4 (18.6% higher, 95% CI, -2.0% to 39.3%; $p = 0.076$), and physicians with conventional resources (20.2% higher, 95% CI, -0.4% to 40.9%; $p = 0.055$).

### Diagnostic Probabilistic Reasoning Cases

We used five cases on primary care topics given to a nationally representative sample of 553 medical practitioners (290 resident physicians, 202 attending physicians, and 61 nurse practitioners or physician assistants) in performing probabilistic reasoning compared with scientific reference probabilities (*20*). As shown in Figure 6 and Table 3, o1-preview performs similarly to GPT-4 in estimating pre-test and post-test probabilities. The exception is the stress test for coronary artery disease, in which o1-preview density is closer to the reference range than models and humans.

### Emergency Room Cases

We compared the ability of o1, 4o, and two attending physicians to provide differential diagnoses across 79 cases from the Beth Israel Deaconess Medical Center, divided into three diagnostic touchpoints (initial ER triage, ER physician, and admission to the medical floor or intensive care unit). Overall, o1 outperformed both 4o and two expert attending physicians, as assessed by two other attending physicians who both were blinded to the source of the differential diagnosis (human or AI model) (Figure 7). Blinding was successful: physician accuracy in guessing AI or Human was 14.8% for one physician (83.5% “Can’t tell”) and 2.7% for the other (84.6% “Can’t tell”). Across each diagnostic touchpoint, o1 either performed better than or on par with the two attending physicians and 4o. Performance differences were especially pronounced at the first diagnostic touchpoint (initial ER triage), where there is the least information available about the patient and the most urgency to make the correct decision.

The o1 model identified the exact or very close diagnosis (Bond scores of 4-5) in 65.8% of cases during the initial ER Triage, 69.6% during the ER physician encounter, and 79.7% at the ICU —surpassing the two physicians (54.4%, 60.8%, 75.9% for Physician 1; 48.1%, 50.6%, 68.4% for Physician 2) at each stage.

## DISCUSSION

We systematically evaluated the medical reasoning abilities of an LLM across six diverse experiments, comparing the model to hundreds of expert physicians. We found consistent superhuman performance in every experiment. Most importantly, the model outperformed expert physicians in real cases utilizing real and unstructured clinical data in an emergency department. These diagnostic touchpoints mirror the high-stakes decisions taken in emergency medicine, where nurses and clinicians make time-sensitive decisions with limited information. Our results showed that both humans, GPT-4o, and o1 all improved their diagnostic abilities as more information was available; however, both LLMs consistently outperformed humans, with the widest margin in low information settings with o1.

The rapid pace of improvement in LLMs has major implications for science and the practice of clinical medicine. While applying AI to assist with clinical decision support is sometimes viewed as a high-risk endeavor (*21*, *22*), greater use of these tools might serve to mitigate the enormous human and financial costs of diagnostic error, delay, and lack of access (*23*, *24*). Our findings suggest the need for prospective trials to evaluate these technologies in real-world patient care settings and prepare for investments for computing infrastructure and design for clinician-AI interaction that can facilitate the integration of AI tools into patient-care workflows. This includes the development of robust monitoring frameworks to oversee the broader implementation of AI clinical decision support systems (*21*).

Our study has several limitations. First, while some of the experiments were originally performed with human-computer interaction, our current study reflects only model performance. Further studies should be done to elucidate how humans and LLMs collaborate together. Second, our study examined only six aspects of clinical reasoning; researchers have identified dozens of other tasks that could be studied which may have even more impact on actual clinical care (*25*). Third, despite large numbers and varieties of cases included in our study which were focused on internal medicine and emergency medicine, it is not representative of broader medical practice which includes multiple subspecialties that require a variety of skill sets such as surgical decisions. There could be varying performance based on diagnosis, patient characteristics, or practice location that are not found in our study. Fourth, while the results of our emergency department experiment appear to have face validity, the task we studied – providing a second opinion at predefined touchpoints – is best thought of as a proof-of-concept. Decisions in the emergency department are often centered around triage, disposition, and immediate management and not diagnostic accuracy.

Overall, our findings show that LLMs now demonstrate superhuman performance in differential diagnosis, diagnostic clinical reasoning, and management reasoning, and are superior in multiple domains compared to prior model generations and even expert human physicians. These same performance gains are seen in providing second opinions in real, unstructured medical cases in the emergency department, where clinicians must act quickly with limited and often missing information.

Over 65 years ago, Ledley and Lusted described the standard for evaluating the diagnostic abilities of artificial intelligence (*1*). The broad challenge they laid out of reasoning over complex clinical case vignettes has now been consistently met. We must now focus on human-computer interaction studies and prospective clinical trials to rigorously assess the potential of artificial intelligence systems to improve clinical practice and patient outcomes.

# METHODS

### Model

The o1-preview model was released on September 12, 2024 ("o1-preview-2024-09-12"), and was accessed through OpenAI's Application Programming Interface (API). The o1 model was released on December 5, 2024, and was accessed using a private Azure API.

### NEJM Clinicopathologic Conference Cases

We selected all 143 diagnostic cases from 2021 to September 2024 (cases including the section "Differential Diagnosis"). There were 70 of these cases, published between 2021 and 2022, that were also evaluated in a prior study of GPT-4 (*12*). For differential diagnosis prediction, we adapted the prompt from the prior study of GPT-4 (Supplement 1A). After prediction of differential diagnoses, we queried the model with "What diagnostic tests would you order next given this differential?" in the same conversation.

Our primary outcomes were differential diagnosis quality and the quality of the suggested testing plan. Differential diagnoses were rated independently by two attending internal medicine physicians (Z.K., A.R), using a previously-developed scoring system called the Bond Score (*26*). Bond Scores range from zero to five (*26*), where five represents a differential list that contains the exact target diagnosis and zero represents a differential list that has no suggestions close to the target diagnosis (Supplement 1B). The quality of the testing plan was scored using a Likert scale from zero to two, by comparing the suggested testing plan to the actual testing plan from the case. A score of two indicates that the LLM-suggested testing plan was nearly identical to the actual testing plan; a score of one indicates that the LLM testing plan would have been helpful or yielded the diagnosis via another test not used in the actual case; zero indicates that the LLM testing plan would be unhelpful. The diagnostic test for seven cases could not be scored because a test plan was not applicable (Supplemental 1C). For both the differential diagnosis and diagnostic test selections, a linear-weighted Cohen's kappa was computed to assess interrater agreement, and discordant scores were reconciled through discussion.

Given that o1-preview has a pretraining end date of October 2023, there is a possibility that published NEJM cases are present in the training data. As a sensitivity analysis, we analyzed the performance of the model before and after this cutoff date to assess the presence of memorization (Supplement 1D).

*Statistical Analysis*

Performance of other LLMs and differential diagnosis generators on NEJM CPCs from previous studies are included in Figure 1 based on prior publications (*12*, *26*–*28*); the set of CPCs each model was evaluated on are not the same. Comparison of o1-preview to a historical control of GPT-4 (Figure 2A) was performed using a McNemar's test of identifying a very close or exact diagnosis (i.e., Bond score 4/5 or 5/5) versus not (i.e., Bond score 0/5, 2/5, or 3/5). The 95% confidence intervals for proportions were computed using a one-sample binomial test. The analysis was performed in R version 4.4.2.

**NEJM Healer Diagnostic Cases**

We used 20 NEJM Healer cases separated into four sections representing sequential stages of clinical data acquisition during an encounter - triage presentation, review of systems, physical exam, and diagnostic tests.

Using prompts adapted from a prior study of GPT-4 (Supplement 2A) (*15*), o1-preview was queried to produce a problem representation, prioritized differential diagnosis, and associated justification. The primary outcome of this study was the quality of clinical reasoning documentation measured by the R-IDEA score. The R-IDEA is a validated 10-point scale for evaluating four core domains of documenting clinical reasoning (Supplement 2B) (*16*). For each case and section (80 responses total), two attending internal medicine physicians (E.B.L and P.J.) rated cases. A linear-weighted Cohen's kappa was computed to assess interrater agreement, and then scores were reconciled by a third internal medicine physician (P.B.).
Our secondary outcome was the identification of "cannot-miss" diagnoses, acknowledging that simple accuracy measures alone do not capture the property that different clinical scenarios and diagnoses may have drastically different severity and impact on patient outcomes. For each case, using only the initial triage presentation data, we used a list of "cannot-miss" defined in a prior study verified by attending three internal medicine physicians (*15*). We captured the number of "cannot-miss" diagnoses included in the o1-preview output for the initial triage presentation differential diagnoses. Two of the 20 cases were excluded because there were no "cannot-miss" diagnoses identified in the prior study (*15*).

*Statistical Analysis*

We compared o1-preview to a historical control of GPT-4, attending physicians, and resident physicians from a previous study (*15*). We performed McNemar's test between o1-preview and each group in achieving a perfect R-IDEA score. The proportion of "cannot-miss" diagnoses included by each model was compared by pairwise t-test with Holm-Bonferroni correction. The analysis was performed in R version 4.4.2.

### ***Grey Matters*** **Management Cases**

The five cases were provided to the o1-preview model, using the same prompt as a prior study (Supplement 3A) (*17*). Two attending internal medicine physicians (E.B.L and P.J.) graded each of the five responses based on rubrics generated by a combination of generalists and subspecialists in the prior study (Supplement 3B) (*17*). We normalized the scoring of all rubrics on a 100 point scale. A linear-weighted Cohen's kappa was computed to assess interrater agreement, and discordant scores were reconciled by a third internal medicine physician (P.B.). The primary outcome was the percentage of total points obtained by o1-preview for each of the five cases.

This outcome was compared to a historical control of GPT-4 alone, humans augmented with GPT-4, and humans augmented with non-LLM conventional resources (e.g., UpToDate, internet search, etc.) from a prior study on the same cases (*17*). The prior study collected five GPT-4 responses to all cases, 176 responses from physicians with GPT-4, and 199 responses from physicians with conventional resources (*17*).

*Statistical Analysis*

We used the same methodology as the prior studies evaluating GPT-4 for management and diagnostic reasoning (*17*, *18*). A linear mixed-effects model was used to compare total percentage points scored by o1-preview to the historical controls: GPT-4, humans with GPT-4, and humans with conventional resources from the prior study. The group was used as the fixed effect. A random intercept was included for the case number, accounting for variability in the difficulty of cases. Another random intercept was used for the interaction of trial number (the number of times a participant answered a case) and individual. GPT-4 and o1-preview were treated as a single individual. GPT-4 was run three times in the prior study and the three attempts were nested under a single individual. P-values are computed for each fixed effect using Satterthwaite's approximation for degrees of freedom. The analysis was performed in R version 4.4.2.

### **Landmark Diagnostic Cases**

We asked o1-preview to read six clinical vignettes, adapted from a landmark study for evaluating computed based diagnostic systems (*19*), and respond with the top three differential diagnoses, three factors that favor or oppose each of these diagnoses, the final most likely diagnosis, and three next diagnostic steps (Supplement 4A). We used the same scoring rubric as the previous study (*18*). Each case was scored across four categories: initial diagnoses (one point each), supporting factors (zero-two points), opposing factors (zero-two points), and final diagnosis (one-two points). Additionally, participants could earn zero-two points for each of up to three recommended next steps in patient evaluation (Supplement 4B). We normalized the scoring of all rubrics on a 100 point scale. Two internal medicine attending physicians (A.R. and Z.K.) graded each of the six o1-preview responses. A linear-weighted Cohen's kappa was computed to assess interrater agreement. Scoring discrepancies were reconciled by a third internal medicine physician (P.B.). A percentage score out of the total points was calculated for each of the six cases. Our primary outcome was the percentage of total points obtained by o1-preview for each of the six cases. This outcome was compared to historical controls from GPT-4 alone, physicians with GPT-4, and physicians with conventional resources recorded in the prior study (*18*).

*Statistical Analysis*

The same method was used as in "*Grey Matters* Management Cases" to compare total percentage points scored by o1-preview to the historical controls using a linear mixed effects model. The analysis was performed in R version 4.4.2.

**Diagnostic Probabilistic Reasoning Cases**

We posed the same questions from a prior study – how the probabilities of a case vignette change in response to positive or negative tests – to the o1-preview model, with 100 model outputs generated for each of the 5 cases (*29*). In the previous study, GPT-4 was run with a default temperature of 1 (a parameter modulating the diversity of the output). o1-preview does not accept a temperature setting using OpenAI's API.

*Statistical Analysis*

Mean absolute error (MAE) and mean absolute percentage error were computed to compare predictions to the reference probabilities collected from the previous literature review (*20*). The analysis was performed in R version 4.4.2.

**Blinded Physician Evaluation of Emergency Room Cases**

We randomly selected 80 cases during a two week period from Beth Israel Deaconess Medical Center in Boston, MA, of patients who presented to the emergency room and were either admitted to the general medicine service or the medical intensive care unit. We were unable to retrieve data for one case, leaving 79 cases. We captured three touchpoints – initial emergency room triage (where the patient is seen by a nurse to determine how quickly they should be seen by a physician), on evaluation by the emergency room physician, and on admission to the medical floor or intensive care unit.

Each touchpoint consisted of unique data drawn directly from the EHR. At triage, nurse-recorded data included sex, age, chief complaint, presumptive diagnosis, triage nurse note, acuity number, means of arrival, and initial vitals. At the end of the emergency department visit, the provider's history of present illness, physical exam, medical decision making, imaging reports, and labs were collected. In many cases the provider was a resident, in which case the attending attestation history of present illness, physical exam, and medical decision making was also captured. Once the patient arrived on the medical floor or medical ICU, the admission chief complaint, history of present illness, physical exam, and assessment and plan were captured. Similarly to the emergency room, many admission notes were completed by residents in which case the attending attestation history of present illness, physical exam, and assessment and plan were also captured. Two attending internal medicine physicians (Z.K., A.R), then reviewed the entire medical record – including follow up visits – for each case independently and recorded the final diagnosis. Disagreements were reconciled in person among the two physicians after the independent review.

The data from the three touchpoints were then provided to GPT-4o and o1 using standardized prompts (Supplement 5A). The same data was provided to two board-certified internal medicine physicians. Using only the information provided in the touchpoints, each physician generated their own second opinion differential (a total of 237 per physician). Because human and LLM differential generation follows different semantic structures, we developed a method for blinding the source of the differentials. We first used a regular expression match to extract individual diagnosis elements from the written responses. We then use a locally-hosted LLM to take each disease name and expand any acronyms used (e.g. “PE” would be converted to “Pulmonary Embolism”). Finally, we converted back to a ranked list, limited to only the top 5 diagnoses. Study member A.H., an ED physician, reviewed 100 pairs of blinded and original differentials to ensure the blinding process captured the same semantic meaning. In 99/100 there was no difference. One error was identified in the blinding process where the differential was incorrectly transformed into a list of follow-up tests. In this case, the differential was manually blinded.

After blinding, Z.K. and A.R. graded blinded LLM and human-generated differentials across the three touchpoints, 945 in total, scoring the differential quality using the Bond score. To assess blinding, Z.K. and A.R. also guessed whether each differential diagnosis was human-generated, LLM-generated, or unsure. Any disagreements were reconciled by discussion. A Cohen’s kappa was calculated to determine concordance.

*Statistical Analysis*

Across the 79 cases and 3 touchpoints, 2 responses were left unanswered by Physician 1 and 1 response was unanswered by o1. We assigned these a Bond score of 0. This resulted in 948 total responses. We then measured the proportion of high-scoring responses, as defined by a reconciled Bond Score of either 4 or 5.

**Figure 1. Performance of Differential Diagnosis Generators and LLMs on *NEJM* Clinicopathologic Case Conferences (CPCs) 2012-2024**

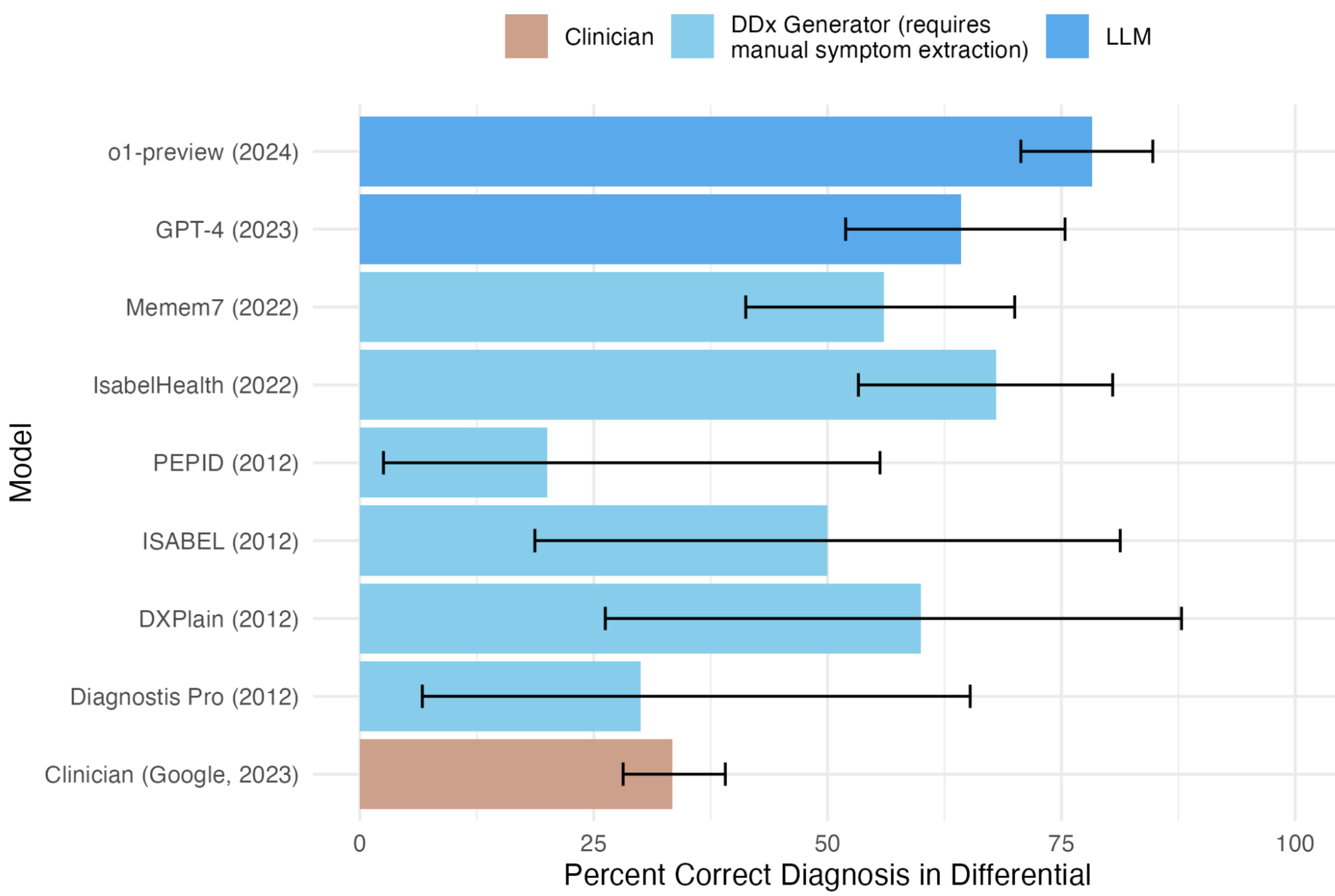

**Figure 1:** Barplot showing the accuracy of including the correct diagnosis in the differential for differential diagnosis (DDx) generators and LLMs on the *NEJM* CPCs, sorted by year. Data for other LLMs or DDx generators was obtained from the literature (*26*) (*28*) (*12*). The 95% confidence intervals were computed using a one-sample binomial test.

## Figure 2. Quality of o1-preview and GPT-4 Differential Diagnosis on NEJM Clinicopathologic Case Conferences (CPCs)

### A. Proportion of Responses Containing the Exact or Very Close Diagnosis for the o1-preview model vs. GPT-4

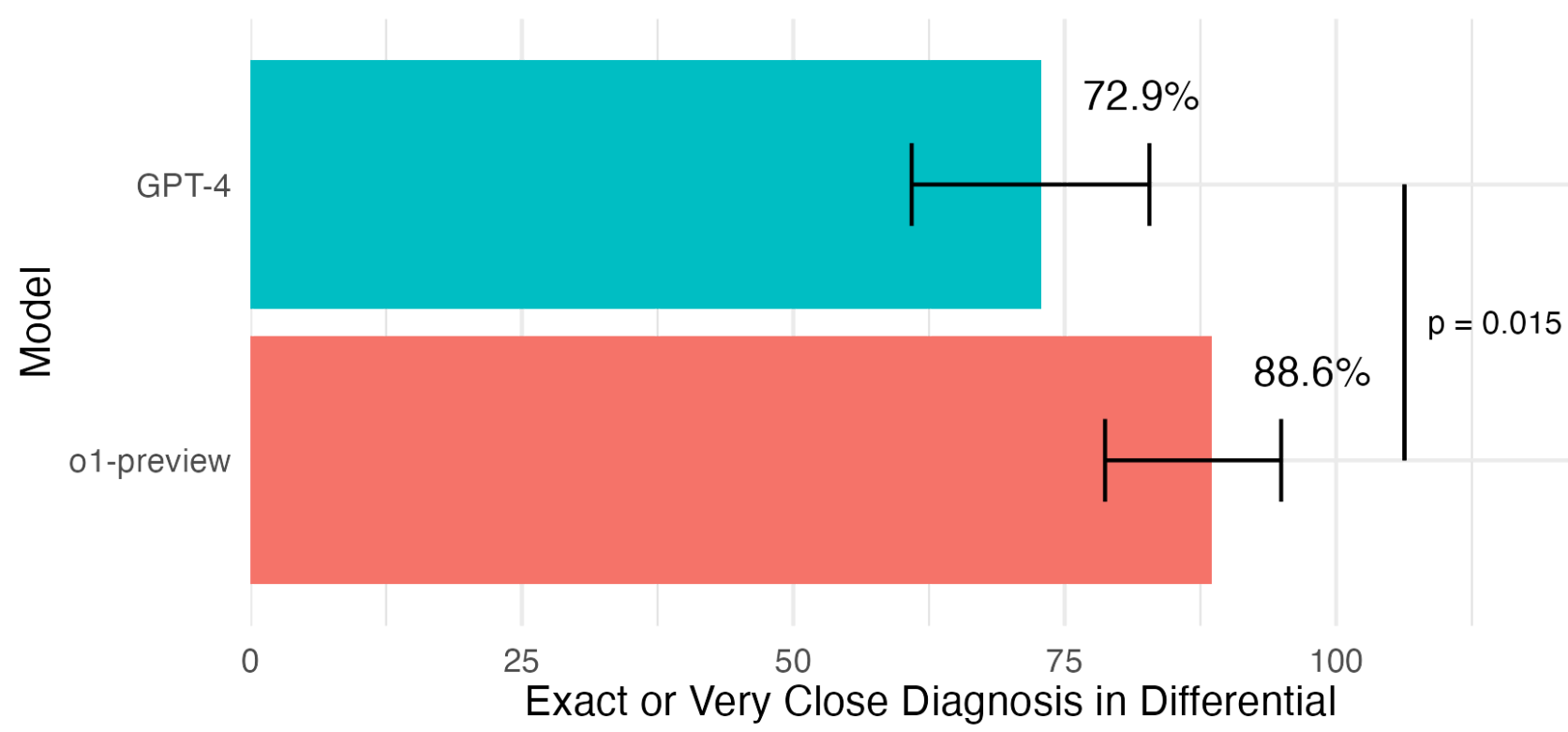

### B. o1-preview Bond Score Distribution on 143 Cases from 2021-2024

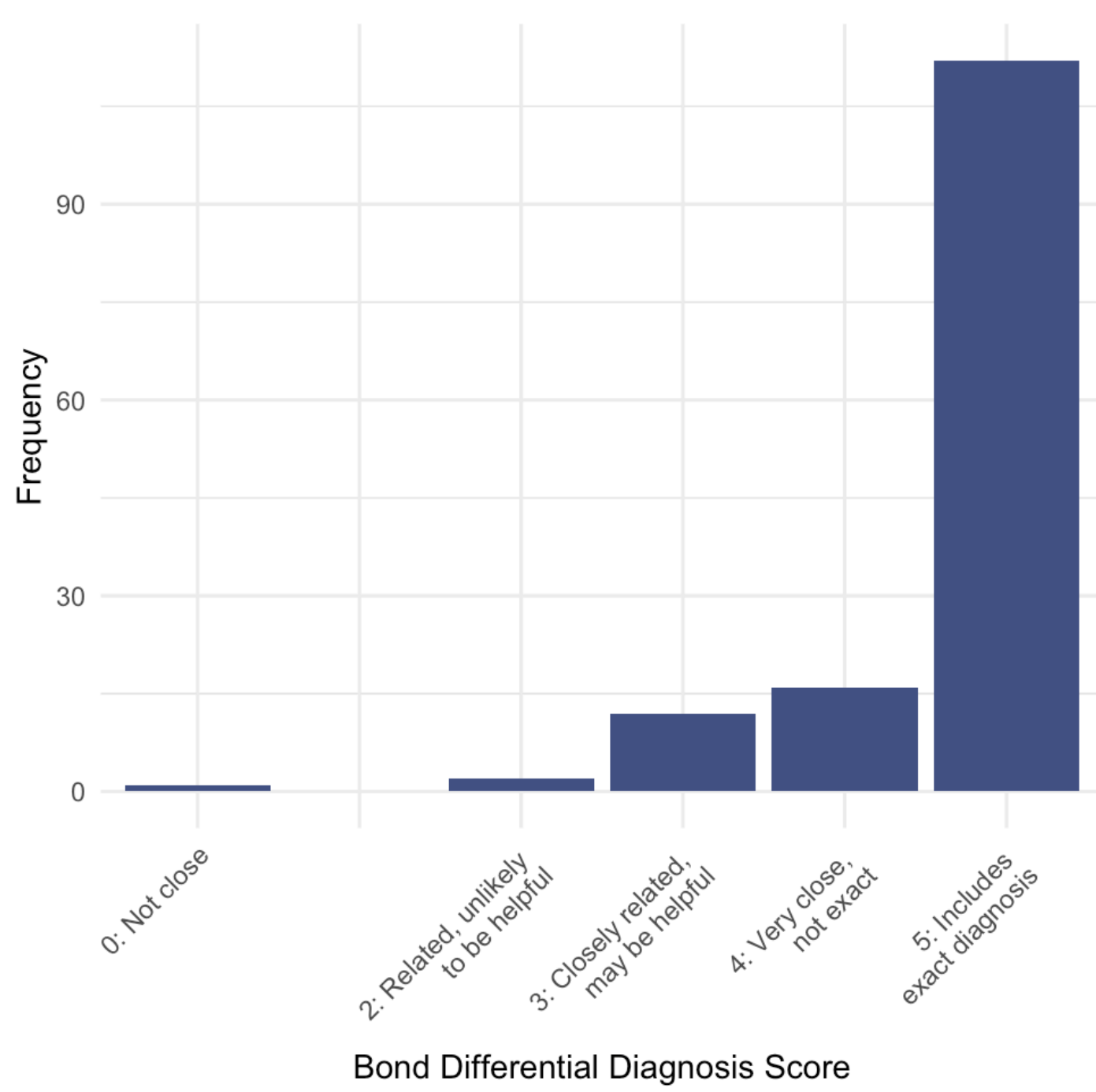

**Figure 2: A.** Comparison of o1-preview with a previous evaluation of GPT-4 in providing the exact or very close diagnosis (Bond scores 4-5) on the same 70 cases. Bars are annotated with the accuracy of each model. 95% confidence intervals were computed using a one-sample binomial test. P-value was computed using McNemar's test. **B.** Histogram of o1-preview performance as measured by the Bond Score on the complete set of 143 cases.

**Figure 3. Quality of o1-preview Diagnostic Test Selection on NEJM Clinicopathologic Case Conferences (CPCs)**

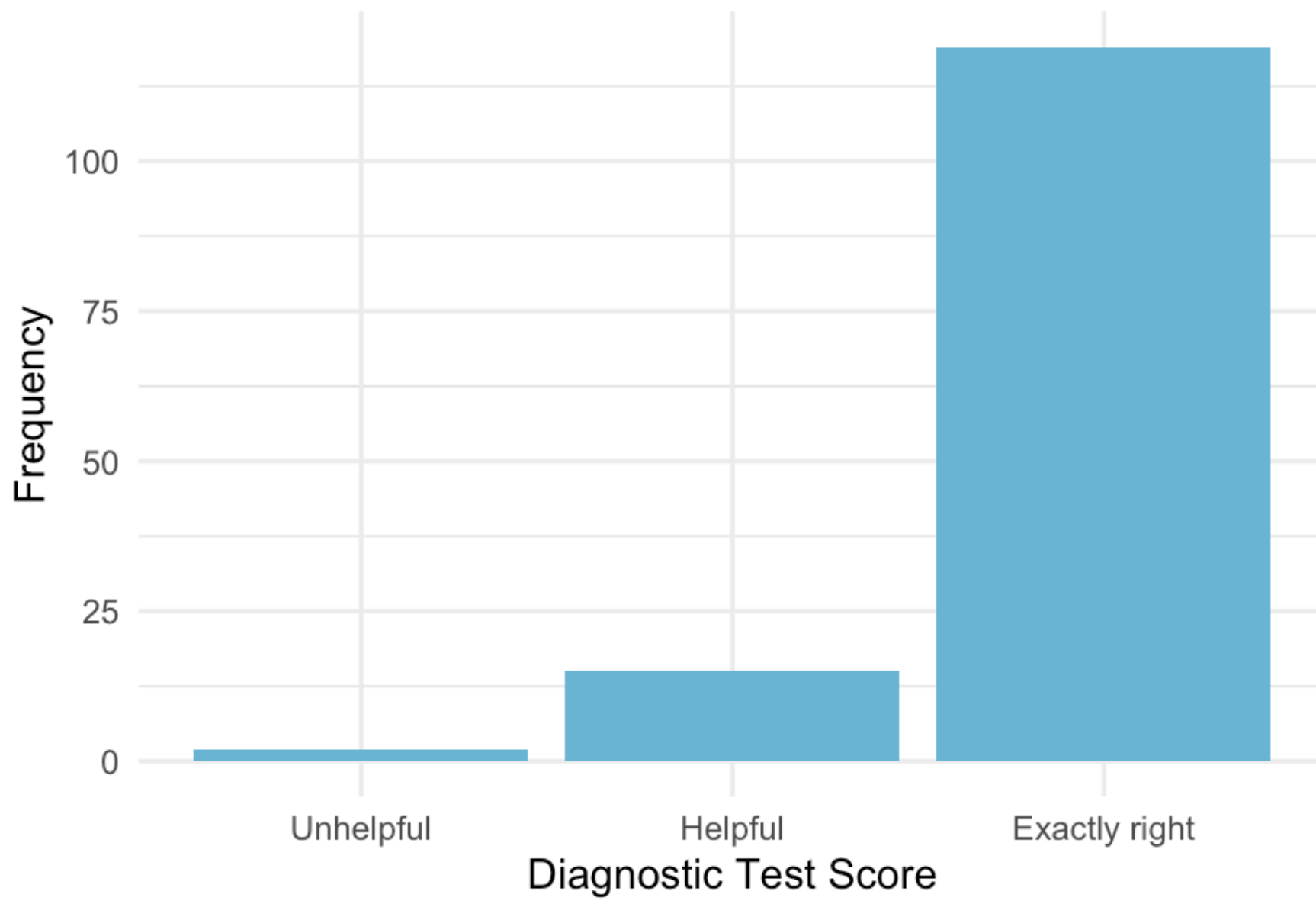

**Figure 3:** Performance of o1-preview in predicting the next diagnostic tests that should be ordered. Performance was measured by two physicians using a likert scale of “Unhelpful,” “Helpful,” and “Exactly right.” We excluded 7 cases from the total case set in which it did not make sense to ask for the next test (Supplement 1B).

## Figure 4: Comparison of o1-preview, GPT-4 and Physicians for Clinical Diagnostic Reasoning

### A. Distribution of R-IDEA Scores on NEJM Healer Cases

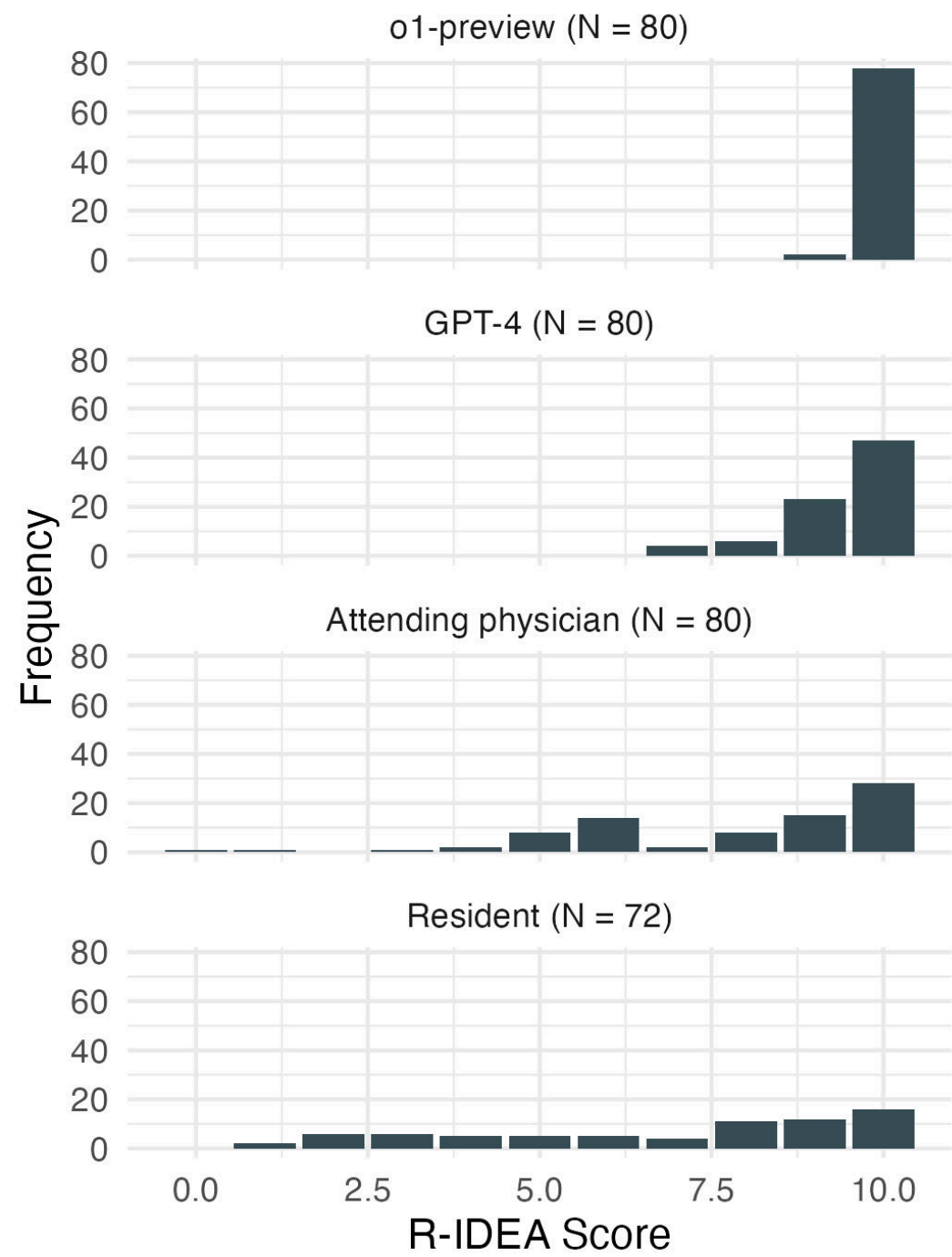

### B. Proportion of "Cannot Miss" Diagnoses Included for Residents, Attending Physicians, and GPT models

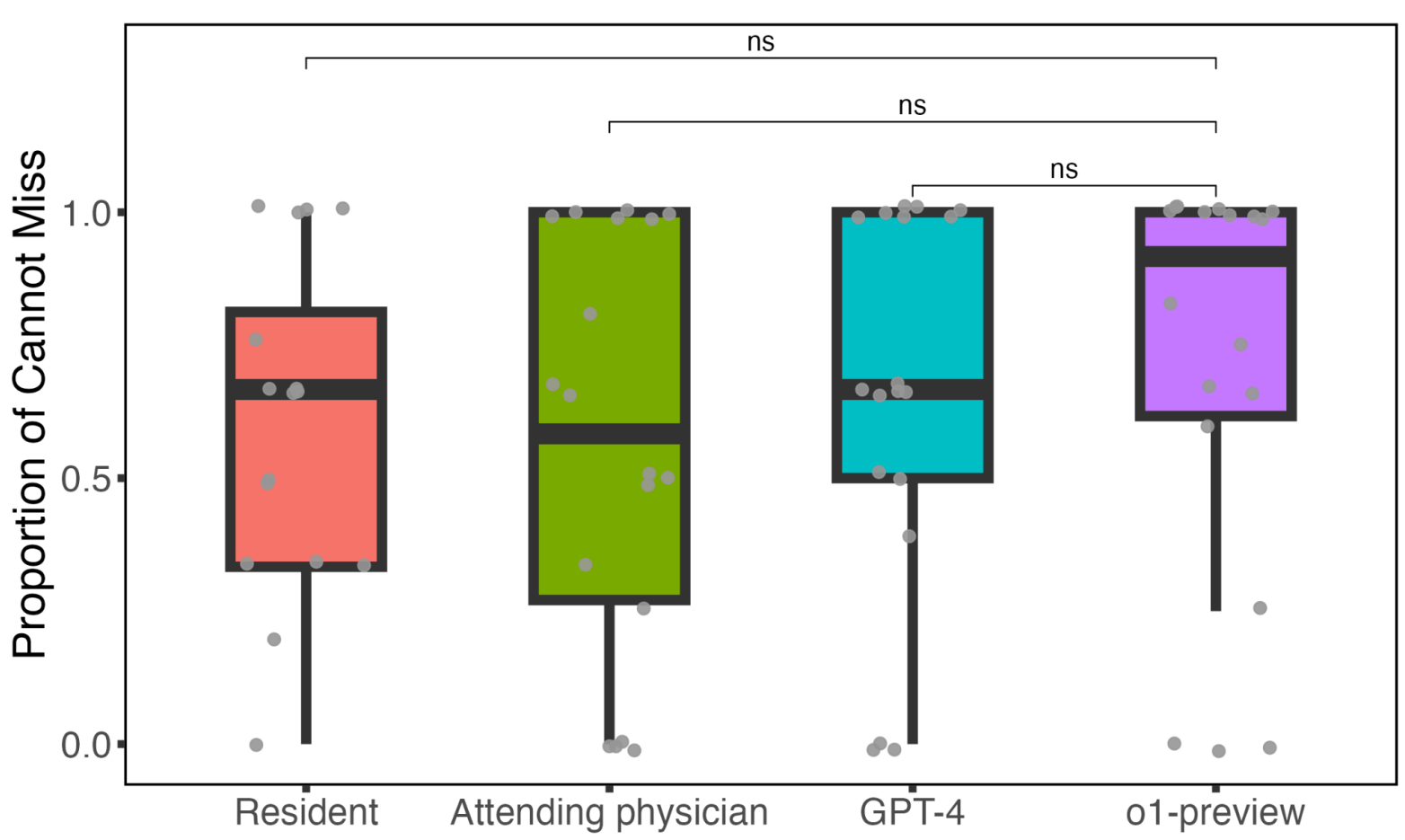

**Figure 4: A.** Distribution of 312 R-IDEA scores stratified by respondents on 20 NEJM Healer cases. **B.** Box plot of the proportion of cannot-miss diagnoses included in differential diagnosis for the initial triage presentation. The total sample size in this figure is 70, with 18 responses from attending physicians, GPT-4 and o1-preview, and 16 responses from residents. Two cases were excluded because the cannot-miss diagnoses could not be identified. Ns: not statistically significant.

## Figure 5: Comparison of o1-preview, GPT-4 and Physicians for Management and Diagnostic Reasoning

### A. Grey Matters Management Cases: o1-preview Management Reasoning Scores Compared to GPT-4 and Physicians

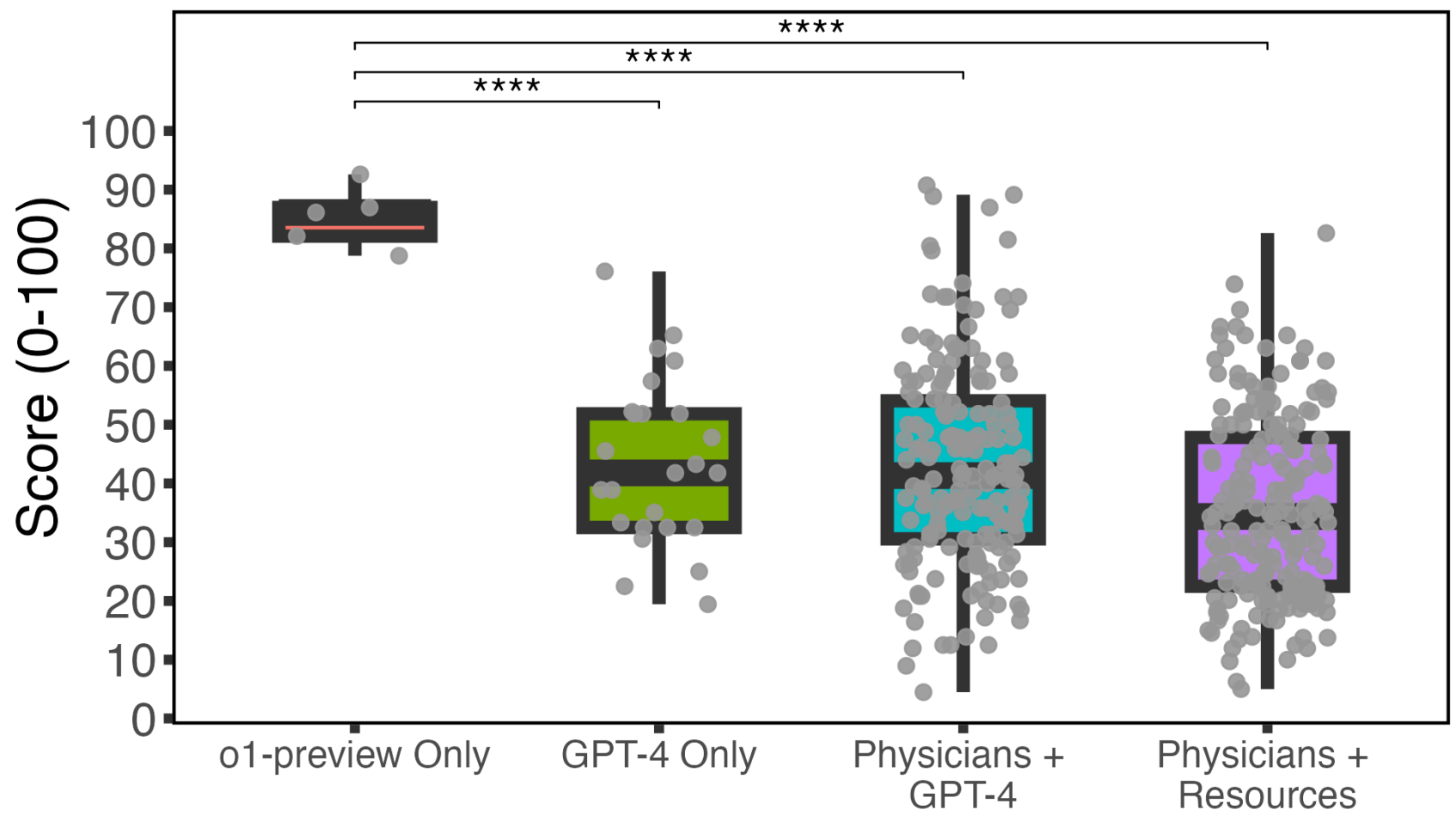

### B. Landmark Diagnostic Cases: o1-preview Diagnostic Reasoning Scores Compared to GPT-4 and Physicians

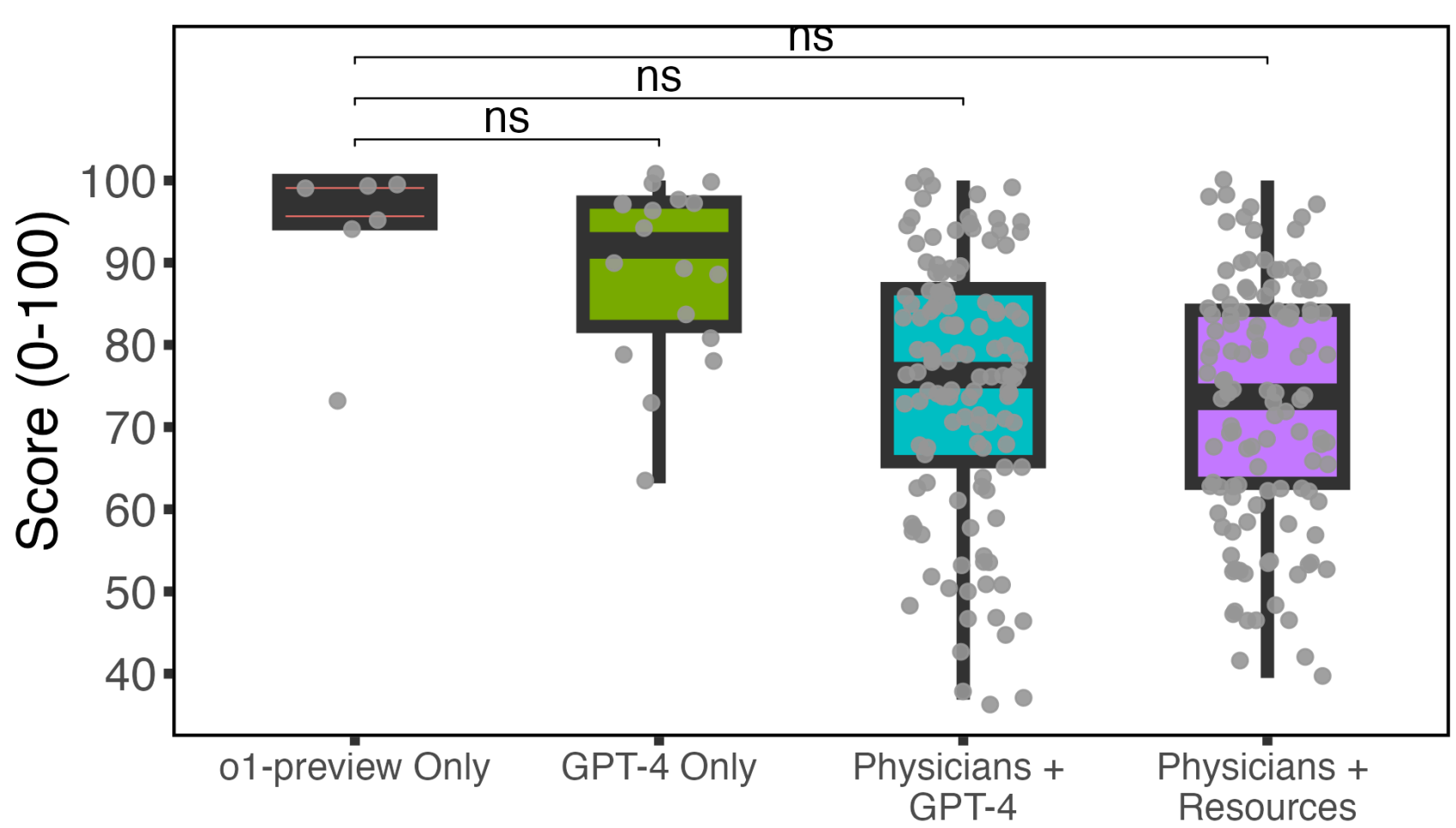

**Figure 5: A.** Box plot of normalized management reasoning points by LLMs and physicians. Five cases were included. We generated one o1-preview response for each case. The prior study collected five GPT-4 responses to each case, 176 responses from physicians with access to GPT-4, and 199 responses from physicians with access to conventional resources. *: p <= 0.05, **: p <= 0.01, ***: p <= 0.001, ****: p <= 0.0001.
**B.** Box plot of normalized diagnostic reasoning points by model and physicians. Six diagnostic challenges were included. We generated one o1-preview response for each case. The prior study collected three GPT-4 responses to all cases, 25 responses from physicians with access to GPT-4, and 25 responses from physicians with access to conventional resources. Ns: not statistically significant.

## Figure 6: Probabilistic Reasoning Before and After Testing by o1-preview

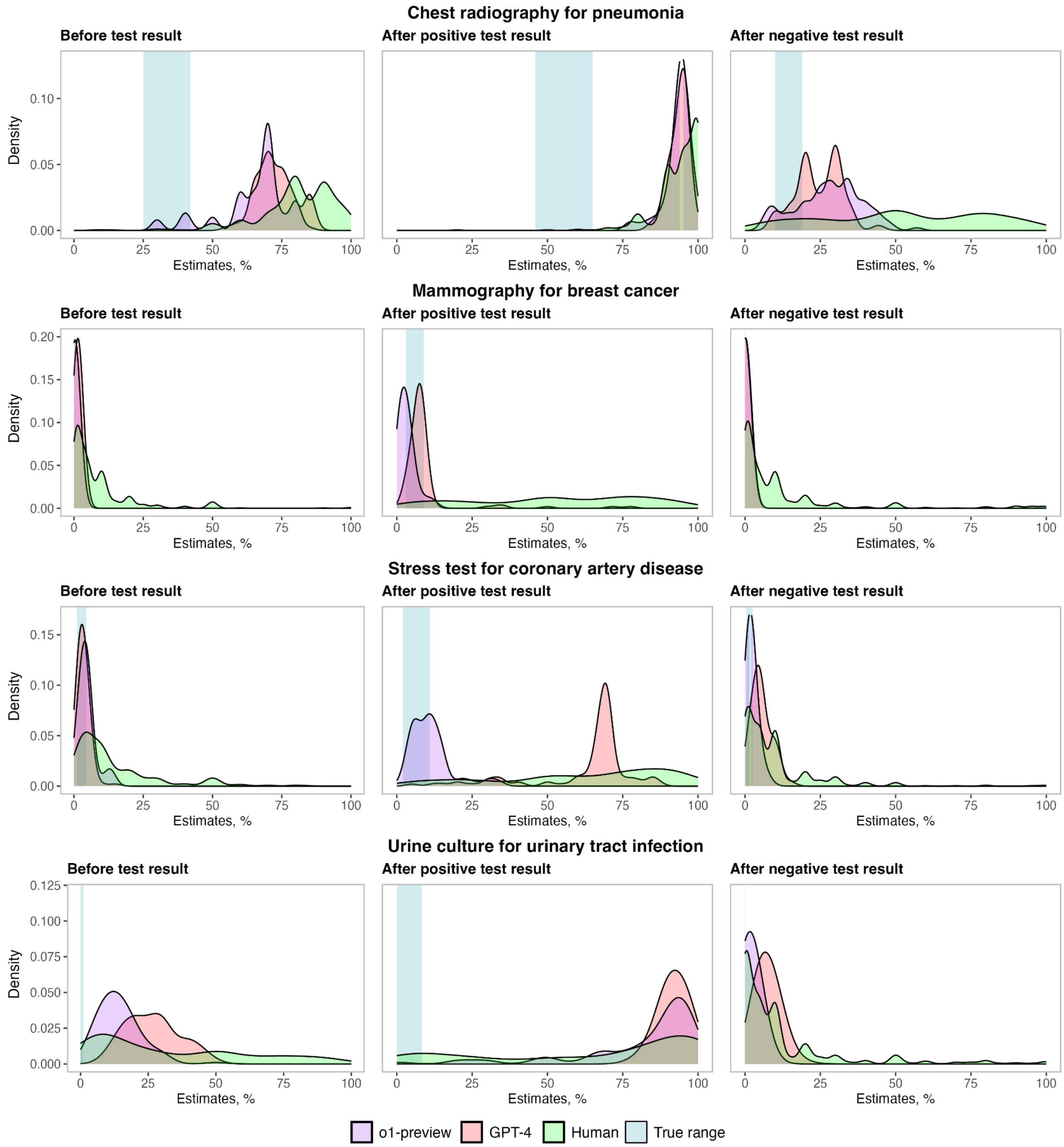

**Figure 6:** Density plots for the distribution of responses by o1-preview, GPT-4 and humans to clinical vignettes asking for (1) the pretest probability of disease, (2) the updated probability after a positive test result, and (3) the updated probability after a negative test result. The shaded blue indicates the reference range based on a review of literature from a prior study.(*20*) Human responses are from 553 medical practitioners (290 resident physicians, 202 attending physicians, and 61 nurse practitioners or physician assistants). 100 predictions were generated by GPT-4 and o1-preview for each question.

**Figure 7: Blinded Assessment of AI and Human Expert Second Opinions on Real Emergency Room Cases**

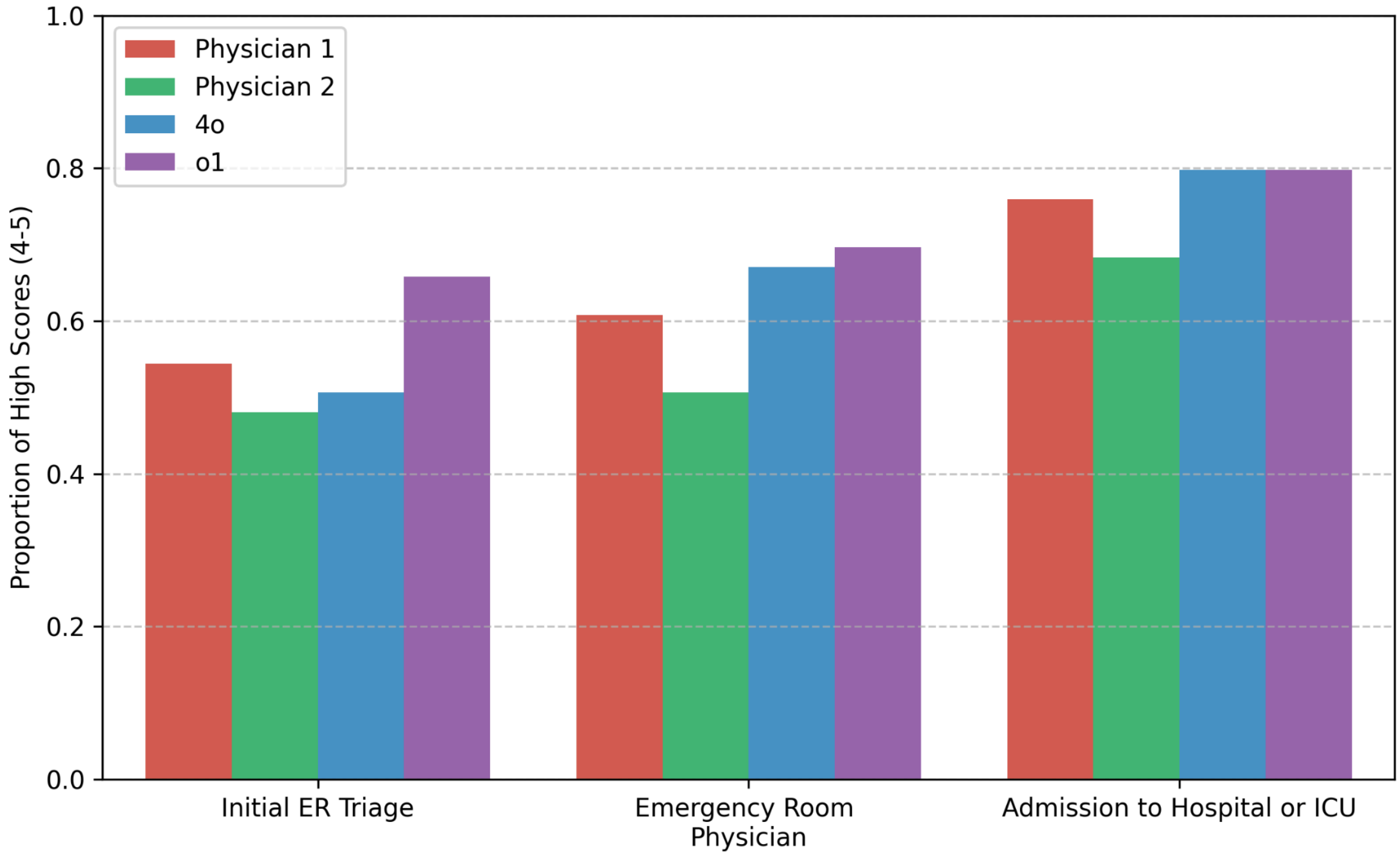

**Figure 7:** Barplot comparing two internal medicine attending physicians, o1, and GPT-4o diagnostic performance on 79 clinical cases at three diagnostic touchpoints (triage in the emergency room, initial evaluation by a physician, and admission to the hospital or intensive care unit). Differential diagnoses were capped at 5 diagnoses for all participants. The source of the differential diagnosis was blinded and scored by two separate attending internal medicine physicians using the Bond scale. The proportion of responses scored 4 or 5 are shown, indicating a response that contains something exact or very close to the true diagnosis.

## Table 1. Examples of o1-preview Correctly Diagnosing Complex Case That GPT-4 Could Not

| Case | Final Diagnosis | GPT-4 Differential | o1-preview Differential |
|---|---|---|---|
| 26-2022 | Langerhans-cell histiocytosis of the lung and liver. | Eosinophilic granulomatosis with polyangiitis (EGPA)<br>Hypereosinophilic syndrome (HES)<br>Eosinophilic pneumonia<br>Parasitic infection (e.g., strongyloidiasis or visceral larva migrans)<br>Disseminated fungal infection (e.g., histoplasmosis or coccidioidomycosis)<br>Lymphoma with eosinophilic infiltration<br>Metastatic cancer with eosinophilic infiltration<br>Sarcoidosis with eosinophilic infiltration<br>Drug-induced eosinophilic syndrome<br><br>Bond Score 0 | **Most Likely Diagnosis:**<br>**Pulmonary Langerhans Cell Histiocytosis (LCH) with Extrapulmonary Involvement**<br>**Differential Diagnosis (Ranked by Probability):**<br>1. **Pulmonary Langerhans Cell Histiocytosis (LCH):**<br>2. **Metastatic Malignancy (e.g., Breast or Thyroid Cancer):**<br>3. **Parasitic Infection (e.g., Echinococcosis):**<br>4. **Hypereosinophilic Syndrome (HES):**<br>5. **Churg-Strauss Syndrome (Eosinophilic Granulomatosis with Polyangiitis):**<br>6. **Sarcoidosis:**<br>7. **Fungal Infections (e.g., Histoplasmosis, Blastomycosis):**<br><br>Bond Score 5 |
| 37-2021 | Anti–melanoma differentiation–associated protein 5 (anti–MDA-5) dermatomyositis. | Coccidioidomycosis (Valley fever)<br>Systemic lupus erythematosus (SLE)<br>Disseminated histoplasmosis<br>Behçet's disease<br>Mixed connective tissue disease<br>Paraneoplastic syndrome<br>Reactive arthritis<br>Sarcoidosis<br>Vasculitis (e.g. granulomatosis with polyangiitis)<br>Parvovirus B19 infection<br><br>Bond Score 0 | **Most Likely Diagnosis:**<br>The most likely diagnosis is **anti-MDA5-positive dermatomyositis** (clinically amyopathic dermatomyositis associated with anti-MDA5 antibodies).<br>**Differential Diagnosis (Ranked by Probability):**<br>1. **Anti-MDA5-Positive Dermatomyositis**:<br>2. **Adult-Onset Still's Disease**:<br>3. **Disseminated Coccidioidomycosis**:<br>4. **Systemic Lupus Erythematosus (SLE)**:<br>5. **Anti-Synthetase Syndrome**:<br>6. **Paraneoplastic Syndrome Associated with Occult Malignancy**:<br>7. **Vasculitis (e.g., ANCA-Associated Vasculitis)**:<br>8. **Sarcoidosis**:<br><br>Bond Score 5 |
| 36-2021 | Erysipelothrix rhusiopathiae infection. | Streptococcus pyogenes cellulitis with possible necrotizing fasciitis<br>Staphylococcus aureus cellulitis with possible necrotizing fasciitis<br>Pasteurella multocida cellulitis from animal exposure<br>Capnocytophaga canimorsus cellulitis from animal exposure<br>Clostridium perfringens gas gangrene or myonecrosis<br>Vibrio vulnificus cellulitis (though less likely due to lack of water exposure)<br>Compartment syndrome secondary to inflammation and swelling<br>Erysipelas caused by Group A Streptococcus<br>Lyme disease-associated cellulitis (unlikely due to rapid onset and severity)<br><br>Bond Score 3 | **Most Likely Diagnosis:**<br>**Cutaneous Anthrax caused by *Bacillus anthracis***<br>**Differential Diagnosis (Ranked by Likelihood):**<br>1. **Cutaneous Anthrax (*Bacillus anthracis*)**<br>2. **Ulceroglandular Tularemia (*Francisella tularensis*)**<br>3. **Erysipeloid Infection (*Erysipelothrix rhusiopathiae*)**<br>4. **Necrotizing Soft Tissue Infection (Group A *Streptococcus*)**<br>5. **Staphylococcal Skin and Soft Tissue Infection (*Staphylococcus aureus*)**<br>6. **Cat-Scratch Disease (*Bartonella henselae*)**<br>7. **Tick-Borne Relapsing Fever (*Borrelia* species)**<br><br>Bond Score 5 |

**Table 1:** Three examples in which o1-preview correctly diagnosed a complex case that GPT-4 could not solve. GPT-4 examples are from a prior study (*12*)

## Table 2. Examples of o1-preview Suggested Testing Plan Compared to Testing Plan Used in the Case

| Case | Case Test Plan | o1-preview Suggested Test Plan | Score |
|---|---|---|---|
| Case 18-2022 | Dr. Selen: Blood and urine samples for the measurement of phosphorus and creatinine levels were obtained while the patient was fasting. The fractional excretion of phosphate was calculated with the following equation: (urine phosphorus level×blood creatinine level×100)÷(blood phosphorus level×urine creatinine level). In the context of hypophosphatemia, a normal fractional excretion of phosphate value is less than 5%. The fractional excretion of phosphate in this patient was 11.6% — a result that suggests that the kidneys were excreting excess phosphate, prompting us to evaluate the blood levels of FGF23.<br>Dr. Yin P. Hung: Intact circulating FGF23 is bioactive. Cleavage of intact FGF23 generates a biologically inactive N-terminal fragment and a biologically inactive C-terminal fragment. There are generally two types of commercial enzyme-linked immunosorbent assays that use antibody conjugates for assessment of FGF23. The iFGF23 assay measures intact FGF23, and the cFGF23 assay provides a composite measurement of both the intact form and the C-terminal fragment of FGF23. The blood level of cFGF23 in this patient was found to be elevated, at 202 reference units (RU) per milliliter (reference value, ≤180), which confirmed the diagnosis of FGF23-dependent hypophosphatemia.<br>Dr. Selen: To determine the source of the elevated level of FGF23, additional imaging studies were obtained.<br>Dr. Balza: Combined gallium-68–dotatate positron-emission tomography–computed tomography was performed (Figure 2). Several sites of abnormal uptake were present in the axial and appendicular skeleton. Uptake was most pronounced in a sclerotic lesion in the left femoral head (the most likely site of tumor in patients with PMTMCT) and was slightly less pronounced in known fractures in the right foot. Other notable sites of uptake included multiple bilateral rib fractures, a sacral fracture, bilateral pubic fractures, a fracture in the right calcaneus, and a lytic lesion in the C7 vertebral body.<br>Diagnostic Imaging Studies.<br>To further characterize the lesion in C7, an MRI of the cervical spine was performed (Figure 2). The lesion had a hyperintense appearance on short-tau inversion recovery images and was slightly hypointense on T1-weighted images. | **1. Laboratory Tests**<br>a. Serum FGF23 Levels<br>b. Fractional Excretion of Phosphate ($FEPO_4$)<br>c. Serum 1,25-Dihydroxyvitamin D Levels<br>d. Comprehensive Metabolic Panel<br>e. Urinalysis and Urine Electrolytes (glucose, amino acids, bicarbonate)<br>f. Parathyroid Hormone (PTH) and PTH-Related Peptide (PTHrP) Levels<br>g. Celiac Disease Serology [  - **Anti-Tissue Transglutaminase (anti-tTG) Antibodies, Anti-Endomysial Antibodies (EMA)**<br>---<br><br>**2. Imaging Studies**<br>- a. Functional Imaging to Locate FGF23-Secreting Tumor:<br>i. 68Ga-DOTATATE PET/CT Scan<br>ii. 18F-Fluorodeoxyglucose (FDG) PET/CT Scan<br>iii. Whole-Body MRI<br>b. Bone Scan (Technetium-99m Bone Scintigraphy)**<br>---<br><br>**3. Genetic Testing**<br>a. Genetic Panels for Hereditary Hypophosphatemic Rickets/Osteomalacia: (identify mutations in genes such as PHEX, FGF23, or DMP1)<br><br>**4. Additional Evaluations**<br>a. Bone Biopsy (if indicated)<br>b. Consultation with Nephrology<br><br>**Summary:**<br>The immediate next steps focus on confirming the presence of elevated FGF23 levels and identifying the source of phosphate wasting. Measuring serum FGF23 is crucial for supporting the diagnosis of Tumor-Induced Osteomalacia. Imaging studies, particularly functional imaging like the 68Ga-DOTATATE PET/CT scan, are essential for localizing the tumor producing FGF23. Simultaneously, tests to evaluate for Fanconi syndrome and other causes of hypophosphatemia should be conducted to either confirm or exclude alternative diagnoses on the differential. Early and accurate identification of the underlying cause will guide appropriate treatment and improve the patient's outcomes. | 2 (completely correct) |
| Case 3-2022 | The first diagnostic test was computed tomographic enterography (CTE), performed after the administration of oral and intravenous contrast material. Coronal images (Figure 1) showed mural thickening and enhancement involving at least 20 cm of distal ileum, a finding consistent with ileitis. The comb sign — dilated, widely spaced, enhancing vasa recta — was present, a finding that reflects bowel inflammation.<br>….When Crohn's disease is suspected, cross-sectional imaging is indicated. Ultrasonography is performed at some centers, but the examination is operator dependent and time consuming. CTE and magnetic resonance enterography (MRE) can be used in the evaluation for Crohn's disease. Both techniques involve the ingestion of a large volume of oral contrast material, as well as the use of intravenous contrast material. It is appropriate to perform CTE as the first cross-sectional examination or when an abscess or perforation is suspected. CTE is fast and easily accessible, and young patients are less likely to be sedated during CTE than during MRE. In comparison, MRE is performed without the use of ionizing radiation and provides a more detailed evaluation of disease activity and bowel motility. Magnetic resonance imaging of the pelvis is the study of choice in the evaluation for suspected perianal disease. | 1. **Gastrointestinal Evaluation:**<br>- **Colonoscopy with Ileoscopy and Biopsies:**<br>- **Esophagogastroduodenoscopy (EGD):**<br>- **Magnetic Resonance Enterography (MRE):**<br>- **Fecal Calprotectin and Lactoferrin Levels:**<br><br>2. **Stool Studies:**<br>- **Comprehensive Stool Culture: (selective media suitable for Yersinia and Campylobacter)<br>- **Stool PCR Panel:**<br>- (bacterial, viral, and parasitic pathogens)<br>- **Stool for Ova and Parasites:**<br>- **Clostridioides difficile Toxin Assay:**<br><br>3. **Serologic and Specific Infectious Disease Tests:**<br>- **Yersinia Serology:**<br>- **Campylobacter Serology:**<br>- **Leptospira Serology (Microscopic Agglutination Test):**<br>- **Lyme Disease Serology:** (Since a test was previously ordered, ensure follow-up on results or consider retesting with ELISA and confirmatory Western blot.)<br>- **Ehrlichia and Anaplasma PCR and Serology:**<br><br>4. **Autoimmune and Rheumatologic Workup:**<br>- **Antinuclear Antibody (ANA) Test:**<br>- **Anti-dsDNA and Anti-Smith Antibodies:**<br>- **Complement Levels (C3, C4):**<br>- **Rheumatoid Factor (RF) and Anti-Citrullinated Protein Antibody (Anti-CCP):**<br>- **HLA-B27 Testing:**<br><br>5. **Laboratory Tests:**<br>- **Complete Blood Count (CBC) with Differential:**<br>- **Erythrocyte Sedimentation Rate (ESR) and C-Reactive Protein (CRP):**<br>- **Comprehensive Metabolic Panel (CMP):**<br>- **Iron Studies and Vitamin B12 Levels:**<br>- **Lipase and Amylase Levels:**<br>- **Urinalysis:** (Detect hematuria or proteinuria)<br><br>6. **Imaging Studies:**<br>- **Abdominal Ultrasound:** (abnormalities such as lymphadenopathy, abscesses, or hepatosplenomegaly).<br>- **Abdominal and Pelvic CT Scan with Contrast:** (identify intestinal inflammation, thickening, or mesenteric adenitis).<br>- **Joint Imaging (X-rays or MRI of Affected Joints):**( joint effusions, erosions, or other signs of arthritis.)<br><br>7. **Additional Tests:**<br>- **Peripheral Blood Smear:** (abnormalities suggestive of leukemia or other hematologic disorders.)<br>- **Bone Marrow Biopsy (if indicated to diagnose leukemia if suspected based on other findings.)<br>- **Serum Complement Levels and Immune Complexes:**<br><br>8. **Consultations:**<br>- **Gastroenterology Consultation:**<br>- **Infectious Disease Consultation:**<br>- **Rheumatology Consultation:**<br>- **Allergy/Immunology Consultation:**<br><br>---<br>**Summary:**<br>A comprehensive diagnostic approach is essential to confirm the most likely diagnosis of Crohn's disease and to exclude other potential causes of the patient's symptoms. Early and accurate diagnosis will guide appropriate management and improve patient outcomes. | 1 |
| Case 30 2023 | Dr. Nina B. Gold: The diagnostic test was measurement of the serum ammonia level. The serum specimen must be obtained from a free-flowing venipuncture without the use of a tourniquet, stored on ice, and sent immediately to the laboratory; deviations from this process may lead to falsely elevated results. This patient's serum ammonia level was 90 μmol per liter (reference range, 12 to 48) and remained elevated on repeat measurements, with a maximum level of 107 μmol per liter. She had encephalopathy during the time that these levels were obtained. The next step was an evaluation for causes of noncirrhotic hyperammonemia, including inherited metabolic disorders and acquired perturbations of the urea cycle.20<br>The plasma amino acid levels obtained during the patient's hospitalization were low, a finding consistent with malnutrition. The free and total plasma carnitine levels, plasma acylcarnitine profile, serum creatine kinase level, and urine organic acid levels obtained after the hospitalization were normal. She was found to have a mildly elevated urine orotic acid level, which can be associated with proximal urea-cycle dysfunction. However, repeat measurements of the plasma amino acid levels have not shown hallmarks of a proximal urea-cycle defect, such as an elevated glutamine level or a low citrulline level (Figure 2). Results of molecular testing for 76 monogenic disorders associated with hyperammonemia were normal, which makes a diagnosis of an inherited metabolic disorder, such as an ornithine transcarbamylase deficiency, unlikely. | 1. **Fecal Alpha-1 Antitrypsin Clearance Test**<br>2. **Technetium-99m-Labeled Albumin Scintigraphy (Radionuclide Imaging)**<br>3. **Lymphangiography or Magnetic Resonance Lymphangiography**<br>4. **Serum Immunoglobulin Levels (IgG, IgA, IgM)**<br>5. **Electrolyte Panel with Focus on Phosphorus, Magnesium, and Potassium Levels**<br>6. **Repeat Electroencephalography (EEG)**<br>7. **Copper and Ceruloplasmin Levels**<br>8. **Comprehensive Nutritional Assessment** [vitamins and minerals, including B vitamins, vitamin D, and fat-soluble vitamins (A, E, K)].<br>9. **Hydrogen/Methane Breath Test for Small Intestinal Bacterial Overgrowth (SIBO)**<br>10. **Jejunal Aspirate and Culture** (SIBO.)<br>11. **Autoimmune Panel Including Anti-Enterocyte and Anti-Goblet Cell Antibodies**<br>12. **Heavy Metal Screening (Blood Levels of Lead, Mercury, Arsenic)**<br>13. **Repeat Gastrointestinal Biopsies with Specialized Staining** (To detect rare conditions not identified on previous biopsies; Special stains (e.g., electron microscopy, immunohistochemistry) might reveal subtle pathologies like mitochondrial disorders or Whipple's disease).<br>14. Serum Prealbumin and Transferrin Levels**<br>15. **MRI of the Brain with Nutritional Protocols** (Certain deficiencies (e.g., vitamin B12, thiamine) can cause characteristic changes visible on MRI).<br>16. **Bone Marrow Biopsy** (for marrow infiltration or hematologic malignancies.)<br>17. Echocardiogram**<br>18. **Skin Biopsy (if there are skin changes)** (dermatologic manifestations of nutritional deficiencies or systemic diseases.)<br>19. **Assessment of Dietary Intake and Absorption Studies** (quantified with tests like D-xylose absorption test.)<br>20. **Genetic Testing for Congenital Disorders of Glycosylation or Lymphatic Development**<br><br>**Prioritization of Tests:**<br>- **First Line:**<br>  - **Fecal Alpha-1 Antitrypsin Clearance**<br>  - **Serum Immunoglobulin Levels**<br>  - **Electrolyte Panel (Phosphorus, Magnesium, Potassium)**<br>  - **Copper and Ceruloplasmin Levels**<br>  - **Comprehensive Nutritional Assessment**<br><br>- **Second Line:**<br>  - **Technetium-99m-Labeled Albumin Scintigraphy**<br>  - **Lymphangiography**<br>  - **Hydrogen/Methane Breath Test**<br>  - **Repeat EEG**<br>  - **Heavy Metal Screening**<br><br>- **Third Line:**<br>  - **Jejunal Aspirate and Culture**<br>  - **Autoimmune Antibody Panel**<br>  - **Specialized Gastrointestinal Biopsies**<br>  - **MRI Brain with Nutritional Protocols** | 0 (incorrect) |

**Table 2:** Three examples of the o1-preview suggested testing plan compared to the testing plan conducted. One example scored a two, indicating that the test was appropriate and nearly identical to the case plan. A score of one indicates that the suggested diagnostics would have been helpful or yielded the diagnosis via another test not used in the case. A score of zero indicates that the suggested diagnostics would be unhelpful. Verbose rationales from o1-preview were abridged by a physician (Z.K.) to better fit in the table.

## Table 3. Probabilistic Reasoning Before and After Testing by o1-preview

| Case | Reference probability range, % | Predicted probability, median (IQR), % | | | MAE (MAPE) | | |
|---|---|---|---|---|---|---|---|
| | | o1-preview (n=100) | GPT-4 (n=100) | Clinician (n=553) | o1-preview (n=100) | GPT-4 (n=100) | Clinician (n=553) |
| **Pneumonia** | | | | | | | |
| Before test | 25-42 | 70 (60-70) | 72 (68.8-78) | 80 (75-90) | 31.6 (94.2) | 39.5 (117.9) | 47.7 (142.4) |
| After positive test result | 46-65 | 94.9 (92.2-95) | 95 (92.3-95) | 95 (90-100) | 37.5 (67.7) | 38.5 (69.4) | 38.5 (69.4) |
| After negative test result | 10-19 | 28.5 (20-34.2) | 25 (20-30) | 50 (30-80) | 14.8 (102.4) | 10.7 (73.5) | 39.9 (275.1) |
| **Breast Cancer** | | | | | | | |
| Before test | 0.2-0.3 | 0.3 (0.1-0.5) | 1.4 (1.4-1.5) | 5 (1-10) | 0.2 (97.9) | 1.2 (466.4) | 8.5 (3385.4) |
| After positive test result | 3-9 | 2.8 (1.2-4) | 7.8 (7-7.8) | 50 (30-80) | 3.5 (59) | 4.4 (74) | 47.6 (792.9) |
| After negative test result | <0.05 | 0.1 (0-0.1) | 0.2 (0.1-0.3) | 5 (1-10) | 0.1 (260.3) | 0.2 (750.6) | 11.3 (45077.4) |
| **Cardiac ischemia** | | | | | | | |
| Before test | 1-4.4 | 5 (2-5) | 2.5 (2-5) | 10 (5-20) | 2.4 (87.3) | 1.3 (47.2) | 11.9 (439.5) |
| After positive test result | 2-11 | 10.9 (6-13.5) | 68.7 (65-70) | 70 (50-90) | 5.7 (87.1) | 56.5 (869.1) | 56.3 (865.9) |
| After negative test result | 0.4-2.5 | 2 (1-2.2) | 5 (3-5.6) | 5 (1-10) | 1 (69.7) | 4 (273.9) | 8.6 (587.9) |
| **Urinary tract infection** | | | | | | | |
| Before test | 0-1 | 15 (10-20) | 25.6 (20-30) | 20 (10-50) | 13.8 (2752) | 26.2 (5241.7) | 32.4 (6472.9) |
| After positive test result | 0-8.3 | 90 (77-95) | 90 (90-95) | 80 (30-95) | 78.2 (1885.5) | 87.8 (2114.9) | 62.2 (1499.6) |
| After negative test result | 0-0.1 | 1 (1-2) | 5 (5-10) | 5 (0-10) | 1.8 (3182.2) | 7.3 (13294.2) | 11.8 (21495) |

**Table 3:** *: $p <= 0.05$, **: $p <= 0.01$, ***: $p <= 0.001$, ****: $p <= 0.0001$

## References

1. R. S. Ledley, L. B. Lusted, Reasoning foundations of medical diagnosis; symbolic logic, probability, and value theory aid our understanding of how physicians reason. *Science* **130**, 9–21 (1959).
2. K. Brodman, A. J. Erdmann Jr, I. Lorge, C. P. Gershenson, H. G. Wolff, The Cornell Medical Index-Health Questionnaire. III. The evaluation of emotional disturbances. *J. Clin. Psychol.* **8**, 119–124 (1952).
3. F. T. de Dombal, D. J. Leaper, J. R. Staniland, A. P. McCann, J. C. Horrocks, Computer-aided diagnosis of acute abdominal pain. *Br. Med. J.* **2**, 9–13 (1972).
4. E. H. Shortliffe, Mycin: A knowledge-based computer program applied to infectious diseases. *Proc. Annu. Symp. Comput. Appl. Med. Care*, 66–69 (1977).
5. E. B. Ing, M. Balas, G. Nassrallah, D. DeAngelis, N. Nijhawan, The Isabel differential diagnosis generator for orbital diagnosis. *Ophthal. Plast. Reconstr. Surg.* **39**, 461–464 (2023).
6. E. L. Burkett, B. R. Todd, A novel use of an electronic differential diagnosis generator in the emergency department setting. *Cureus* **15**, e34211 (2023).
7. H. Nori, N. Usuyama, N. King, S. McKinney, X. Fernandes, S. Zhang, E. Horvitz, From Medprompt to o1: Exploration of run-time strategies for medical challenge problems and beyond. *arXiv* (2024).
8. OpenAI o1 System Card. https://openai.com/index/openai-o1-system-card/.
9. P. Lee, S. Bubeck, J. Petro, Benefits, limits, and risks of GPT-4 as an AI chatbot for medicine. *N. Engl. J. Med.* **388**, 1233–1239 (2023).
10. S. Bubeck, V. Chandrasekaran, R. Eldan, J. Gehrke, E. Horvitz, E. Kamar, P. Lee, Y. T. Lee, Y. Li, S. Lundberg, H. Nori, H. Palangi, M. T. Ribeiro, Y. Zhang, Sparks of Artificial General Intelligence: Early experiments with GPT-4, *arXiv [cs.CL]* (2023). http://arxiv.org/abs/2303.12712.
11. S. Johri, J. Jeong, B. A. Tran, D. I. Schlessinger, S. Wongvibulsin, L. A. Barnes, H.-Y. Zhou, Z. R. Cai, E. M. Van Allen, D. Kim, R. Daneshjou, P. Rajpurkar, An evaluation framework for clinical use of large language models in patient interaction tasks. *Nat. Med.* **31**, 77–86 (2025).
12. Z. Kanjee, B. Crowe, A. Rodman, Accuracy of a Generative Artificial Intelligence Model in a Complex Diagnostic Challenge. *JAMA* **330**, 78–80 (2023).
13. A. Rodman, L. Zwaan, A. Olson, A. K. Manrai, When it comes to benchmarks, humans are the only way. *NEJM AI* **2**, AIe2500143 (2025).
14. R.-E. E. Abdulnour, A. S. Parsons, D. Muller, J. Drazen, E. J. Rubin, J. Rencic, Deliberate practice at the virtual bedside to improve clinical reasoning. *N. Engl. J. Med.* **386**, 1946–1947 (2022).

15. S. Cabral, D. Restrepo, Z. Kanjee, P. Wilson, B. Crowe, R.-E. Abdulnour, A. Rodman, Clinical Reasoning of a Generative Artificial Intelligence Model Compared With Physicians. *JAMA Intern. Med.* **184**, 581–583 (2024).

16. V. Schaye, L. Miller, D. Kudlowitz, J. Chun, J. Burk-Rafel, P. Cocks, B. Guzman, Y. Aphinyanaphongs, M. Marin, Development of a Clinical Reasoning Documentation Assessment Tool for Resident and Fellow Admission Notes: a Shared Mental Model for Feedback. *J. Gen. Intern. Med.* **37**, 507–512 (2022).

17. E. Goh, R. Gallo, E. Strong, Y. Weng, H. Kerman, J. Freed, J. A. Cool, Z. Kanjee, K. P. Lane, A. S. Parsons, N. Ahuja, E. Horvitz, D. Yang, A. Milstein, A. P. J. Olson, J. Hom, J. H. Chen, A. Rodman, Large language model influence on management reasoning: A randomized controlled trial. *medRxiv*, doi: 10.1101/2024.08.05.24311485 (2024).

18. E. Goh, R. Gallo, J. Hom, E. Strong, Y. Weng, H. Kerman, J. A. Cool, Z. Kanjee, A. S. Parsons, N. Ahuja, E. Horvitz, D. Yang, A. Milstein, A. P. J. Olson, A. Rodman, J. H. Chen, Large language model influence on diagnostic reasoning: A randomized clinical trial: A randomized clinical trial. *JAMA Netw. Open* **7**, e2440969 (2024).

19. E. S. Berner, G. D. Webster, A. A. Shugerman, J. R. Jackson, J. Algina, A. L. Baker, E. V. Ball, C. G. Cobbs, V. W. Dennis, E. P. Frenkel, Performance of four computer-based diagnostic systems. *N. Engl. J. Med.* **330**, 1792–1796 (1994).

20. D. J. Morgan, L. Pineles, J. Owczarzak, L. Magder, L. Scherer, J. P. Brown, C. Pfeiffer, C. Terndrup, L. Leykum, D. Feldstein, A. Foy, D. Stevens, C. Koch, M. Masnick, S. Weisenberg, D. Korenstein, Accuracy of practitioner estimates of probability of diagnosis before and after testing. *JAMA Intern. Med.* **181**, 747–755 (2021).

21. R. M. Ratwani, D. W. Bates, D. C. Classen, Patient safety and artificial intelligence in clinical care. *JAMA Health Forum* **5**, e235514 (2024).

22. Q. Jin, F. Chen, Y. Zhou, Z. Xu, J. M. Cheung, R. Chen, R. M. Summers, J. F. Rousseau, P. Ni, M. J. Landsman, S. L. Baxter, S. J. Al'Aref, Y. Li, A. Chen, J. A. Brejt, M. F. Chiang, Y. Peng, Z. Lu, Hidden flaws behind expert-level accuracy of multimodal GPT-4 vision in medicine. *NPJ Digit. Med.* **7**, 190 (2024).

23. D. E. Newman-Toker, S. M. Peterson, S. Badihian, A. Hassoon, N. Nassery, D. Parizadeh, L. M. Wilson, Y. Jia, R. Omron, S. Tharmarajah, Others, Diagnostic errors in the emergency department: a systematic review. (2023).

24. A. D. Auerbach, T. M. Lee, C. C. Hubbard, S. R. Ranji, K. Raffel, G. Valdes, J. Boscardin, A. K. Dalal, A. Harris, E. Flynn, J. L. Schnipper, UPSIDE Research Group, Diagnostic errors in hospitalized adults who died or were transferred to intensive care. *JAMA Intern. Med.* **184**, 164–173 (2024).

25. M. Goldszmidt, J. P. Minda, G. Bordage, Developing a unified list of physicians' reasoning tasks during clinical encounters. *Acad. Med.* **88**, 390–394 (2013).

26. W. F. Bond, L. M. Schwartz, K. R. Weaver, D. Levick, M. Giuliano, M. L. Graber, Differential diagnosis generators: an evaluation of currently available computer programs. *J. Gen. Intern. Med.* **27**, 213–219 (2012).

27. D. McDuff, M. Schaekermann, T. Tu, A. Palepu, A. Wang, J. Garrison, K. Singhal, Y. Sharma, S. Azizi, K. Kulkarni, L. Hou, Y. Cheng, Y. Liu, S. Sara Mahdavi, S. Prakash, A. Pathak, C. Semturs, S. Patel, D. R. Webster, E. Dominowska, J. Gottweis, J. Barral, K. Chou, G. S. Corrado, Y. Matias, J. Sunshine, A. Karthikesalingam, V. Natarajan, Towards Accurate Differential Diagnosis with Large Language Models, *arXiv [cs.CY]* (2023). http://arxiv.org/abs/2312.00164.

28. P. Fritz, A. Kleinhans, R. Raoufi, A. Sediqi, N. Schmid, S. Schricker, M. Schanz, C. Fritz-Kuisle, P. Dalquen, H. Firooz, G. Stauch, M. D. Alscher, Evaluation of medical decision support systems (DDX generators) using real medical cases of varying complexity and origin. *BMC Med. Inform. Decis. Mak.* **22**, 254 (2022).

29. A. Rodman, T. A. Buckley, A. K. Manrai, D. J. Morgan, Artificial intelligence vs clinician performance in estimating probabilities of diagnoses before and after testing. *JAMA Netw. Open* **6**, e2347075 (2023).